# Coordinating robotized construction using advanced robotic simulation: The case of collaborative brick wall assembly

Mohammad Reza Kolani, Stavros Nousias, André Borrmann
Technical University of Munich, Germany
mohammadreza.kolani@tum.de

**Abstract.** Utilizing robotic systems in the construction industry is gaining popularity due to their build time, precision, and efficiency. In this paper, we introduce a system that allows the coordination of multiple manipulator robots for construction activities. As a case study, we chose robotic brick wall assembly. By utilizing a multi-robot system where arm manipulators collaborate with each other, the entirety of a potentially long wall can be assembled simultaneously. However, the reduction of overall bricklaying time is dependent on the minimization of time required for each individual manipulator. In this paper, we execute the simulation with various placements of material and the robot's base, as well as different robot configurations, to determine the optimal position of the robot and material and the best configuration for the robot. The simulation results provide users with insights into how to find the best placement of robots and raw materials for brick wall assembly.

## 1. Introduction

The construction industry utilizes individualized production lines that are frequently labor-intensive, manual, and involve on-site activities (Naoum, 2016). Consequently, these operations often struggle to meet the increasing demands for rapid and cost-effective methods. Nevertheless, the current construction industry is facing a significant change due to advancements in automation and robotics. Utilizing robotic systems can automate human tasks and speed up construction activities in various operations (Bock & Linner, 2015), such as bricklaying. Collaborative robotic systems in the construction industry hold promise for overcoming these limitations by introducing precision, efficiency, and scalability to construction activities.

An industrial arm robot has high flexibility, but it restricts the task size that can be manipulated due to limitations in the robot's workspace. This issue can be avoided by using larger manipulators; however, investing in larger manipulators is not always financially viable (Zhang et al., 2021). Moreover, moving the robot placement based on the wall geometries is another option; however, relocating the arm robot is time-consuming and not an efficient solution, indicating that utilizing a single manipulator is not a time-efficient approach for building large-scale parts (Bhatt, Kulkarni, et al., 2022). Furthermore, another option is to use arm robots with rails to cover larger working spaces; however, rail installation is cumbersome and requires extra effort, and in some cases, it is impossible due to the unstructured terrain and dynamic environment in construction sites. On the other hand, utilizing a collaborative robotic system and minimizing the time for an individual arm robot can lead to a decrease in the total time required for task completion.

This paper presents a novel system to generate multi-robot placements for different parts of a wall, given certain wall patterns and dimensions. Furthermore, the system proposed the robots' placement and raw material placement to minimize the time spent bricklaying.

The rest of the paper is organized as follows: Section 2 presents the related works; Section 3 is dedicated to the methodology; and Section 4 presents the case study. Section 5 presents the simulation setup and experimental evaluation, while Section 6 concludes our work.

## 2. Related works

Several attempts have been made to automate robotic construction assembly, including the assembly of wooden structures using robotic systems (Koerner-Al-Rawi et al., 2020; Naboni et al., 2021); component transfer using automated tower cranes (Lee et al., 2009); automated positioning and installation robots for steel beam (Nam et al., 2007); the cooperative robotic assembly method for brick vault construction (Parascho et al., 2020); enhancing the precision of the manipulator by determining the optimal positioning of the robot for additive manufacturing purposes (Bhatt et al., 2021); optimization-driven algorithm for placing a single robot that evaluates the velocity and forces of the end-effector tool, as well as conducts a reachability (Malhan et al., 2019); optimization robot placement developed by (Spensieri et al., 2016) aims to enhance the efficiency of automotive assembly and minimize cycle time. Moreover, (Slepicka et al., 2022) proposed Fabrication Information Modelling (FIM) methodology with which the information of a digital building model can be detailed, component by component to transfer to arm robot for execution. Furthermore, Shen et al. (2029) proposed a methodology for additive manufacturing based on a multi-robot system and achieved 73% faster printing time compared to a single-robot printing system. Zhu et al. (2021) investigated a component-oriented robot construction approach. By applying the smart construction object (SCO) paradigm, diverse construction tasks are allocated to robots by assigning states and requirements to the components to drive robots for the assembly of prefabricated housing (Zhu et al., 2021). However, the current literature does not address the combined material and robotic arm placement problems, which is required for efficient robotized bricklaying.

## 3. Methodology

### 3.1 Multi-robot coordination

We propose a methodology for coordinating multi-robot construction tasks. In this methodology, the number of robots required to complete the task is determined based on the task's structure and pattern, the robots' availability, and the robots' working space size. Subsequently, the robots need to be placed in positions in an arrangement that ensures their workspace covers all parts of the task, Figure 1 (a). Furthermore, the robots' placement should satisfy the required working space for task execution while minimizing the shared space to avoid collisions between them. The system divides a task into distinct zones, each corresponding to a specific workspace for individual manipulators. This segmentation is instrumental in defining subtasks tailored to the robots' capabilities and workspace size. Considering each robot's unique working space, the construction task decomposes into several subtasks. Furthermore, subtasks are allocated to the robots to ensure optimal utilization of material and efficient completion of the process. An essential aspect of the methodology is to minimize the task duration for a single manipulator by proposing the best placements of materials and robots. The objective of reducing the time required for a manipulator is to minimize the overall time spent on task completion within a multi-robot system.

Multi-robot system is more versatile and economical as it can be used for different tasks; however, utilizing several arm robots to perform construction tasks presents various challenges. By decomposing the operation into multiple subtasks for the robots, certain parts of the task can be successfully built by the first agent, while another segment of the task can be built by another robot simultaneously. Furthermore, the robots' relative positions can increase or decrease their flexibility and reachability. For example, if the manipulator reachability regions have significant overlap, the multi-robot system provides high flexibility, as shown in Figure 1 (b). This means that if one of the manipulators fails, the other one can take over its task.

However, this limits the size of the task (e.g., a wall) that can be built in the multi-robot system, or it needs more arm robots for a specific size, which is not cost effective (Bhatt, Nycz, & Gupta, 2022). Furthermore, high overlap increases the possibility of collisions between robots and necessitates a complex task-planning system to avoid such collisions. Therefore, minimizing the shared space between robots can provide us with less complicated task planning, Figure 1 (c). Therefore, it is essential to investigate the optimal placement of robots to enhance their reachability and minimize collision probability.

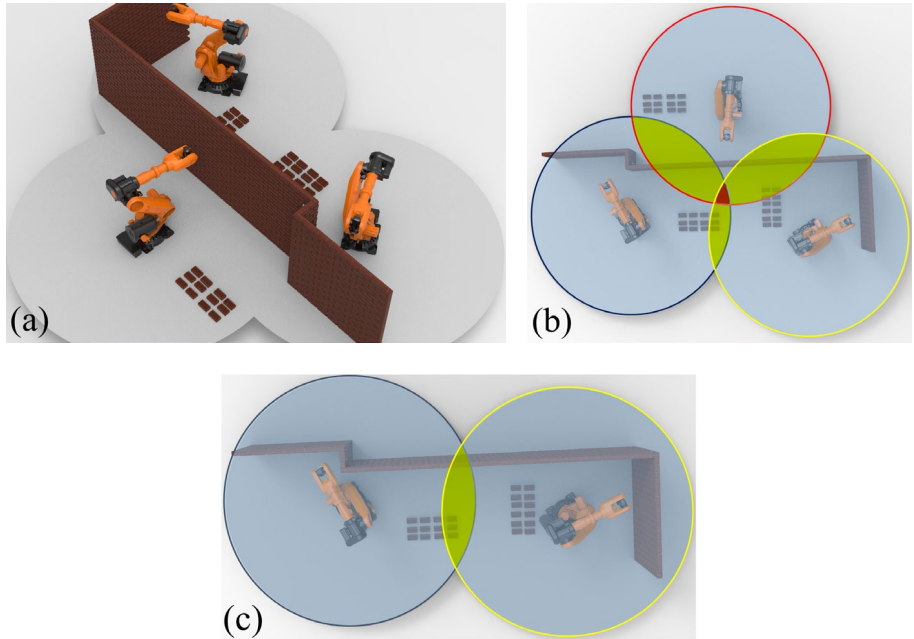

Figure 1: (a) multi-robot coordination for bricklaying case with different shared spaces. (b) covering the wall by three robots (high flexibility and collision probability), (c) covering the wall by two robots (lower flexibility and collision probability)

The optimal placement of the material (bricks) and the robots will emphasize the potential benefits of this approach, such as a reduction in build time, and improved manipulator reachability. In addition, it will reduce the probability of collisions and avoid unnecessary time spent calculating collision avoidance paths. This work has three contributions. (1) It provides novel path decomposition and multi-robot placement algorithms for performing construction activities. (2) providing the best robot's base placement to improve efficiency and operation time. (3) It also provides the best material placement to decrease the operation time.

### 3.2 Simulation framework for construction robotics

We utilize a framework that allows us to plan the deployment of robotic systems for construction tasks. The framework uses existing software and tools for robots' simulation. It relies on the incorporation of the Robot Operating System (ROS[1]) as the underlying framework that facilitates different communication types to connect several nodes in real-time. MoveIt, a powerful robotic framework, stands as a component for motion planning and computing the inverse kinematic of robots for task planning and execution. Furthermore, for visualization purposes, we used Rviz. The integration of ROS, Rviz, and motion planning is the basis of this framework, providing a comprehensive and versatile framework for simulating, planning, and executing using arm manipulators.

---

[1] https://docs.ros.org/en/humble/index.html

## 4. Case study

In this study, we are considering a dry bricklaying case with collaborative robotic arms. The primary objective is to present a system that enables efficient bricklaying operations by determining the optimal number of robots based on the wall's pattern and the robots' working space size. Furthermore, optimizing the initial positioning of the material and robots would enhance the operation's efficiency. The system divides the wall into several zones, with each zone allocated to one arm robot to enable coordinated cooperation (Figure 1). Furthermore, the approach assigns brick positions in each zone to minimize bricklaying time. This approach speeds up the bricklaying by utilizing automated planning and optimization of the placement of the robots and bricks.

The system based on robots sharing workspaces divides the wall into safe zones (blue) and danger zones (yellow and red), as shown in Figure 1 (a-b). The yellow zones are available to a pair of robots, whereas the red zone is accessible to all robots. Therefore, the system provides sequences so that two or more robots are not allowed to work in these zones simultaneously. Considering those zones and our rule that only one robot should operate in a dangerous zone, provide safe multi-robot coordination.

The overall time of bricklaying depends on the number of robots performing the task simultaneously. Since each robot has a specific working space, the number of robots has a direct effect on the total time spent bricklaying. Moreover, for individual robots, the time of picking and placing the bricks depends on two main parts: time for planning ($t_p$) and time for execution ($t_e$). Time for planning means the amount of time that a solver needs to calculate the inverse kinematic for each stage of the activity. The time for execution depends on the robot configuration, the distance of the bricks' initial position to the robot position, and the bricks target positions.

### 4.1 Problem formulation

We consider a set of 7 Degrees Of Freedom (DOF) arm manipulators (Panda) $M = \{M_1, M_1, \ldots, M_n\}$ who are tasked to assemble a brick wall ($n = 3$ in our examples) as shown in Figure 1 (a). Each manipulator is equipped with a gripper and is capable of picking and placing a brick. To perform efficient bricklaying by using this multi-robot system, we need to achieve the following goals:

1- Reduce the assembly time: Assembly time is the total time required by the set of manipulators to build the part completely, $T_{total} = T_1 + T_2 + T_3$. We need to decompose the wall and assign each part to each manipulator. By doing so, the manipulators can simultaneously assemble the part in the minimum amount of time.

2- Find the best robot position: Robot position has a direct impact on the time of execution for an individual robot. For instance, placing a brick takes longer for each manipulator if there is a big distance between the robot and the brick's target position.

3- Find the best material (bricks) position: The duration of bricklaying is directly influenced by the initial position of the bricks for each arm manipulator.

4- Optimal robot trajectory: 7 DOF arm manipulator can consist of multiple Inverse Kinematic (IK) solutions for each stage of the task for the same end effector pose (Spong et al., 2020). The reachability of the manipulator is formed by the combination of IK solutions from various IK families (Bhatt, Nycz, & Gupta, 2022). Ideally, the manipulator should follow the easiest and fastest IK solution while picking and placing bricks.

However, it is not always possible because of joint limits and collision avoidance. Since we are using MoveIt to calculate the IK for every stage and MoveIt handles the IK by utilizing several solvers and optimization algorithms, we do not discuss trajectory planning as it is not within the scope of this paper.

## 4.2 Approach

Our approach consists of two processing steps: decomposing the wall into a set of continuous segments and finding the best positions of the robots' base and material. During the decomposition, it tries to ensure that the decomposed segment sets have equal execution time and there are no gaps between them. This means that the manipulators will have less idle time when simultaneously executing the allocated segment sets. Thus, the wall decomposition module's goal is to reduce the time spent bricklaying. The wall decomposition modules output the $n$ decomposed set of segments $L_1, L_2, ..., L_n$ which is taken as an input to the multi-robot placement module. Finally, the multi-robot placement module returns the placement locations for the $n$ manipulators.

### A. Wall decomposition

Since in our case the number of robots is three, therefore:

$$L_{wall} = L_1 + L_2 + L_3 \qquad (1)$$

Since we used the same robots as our multi-robot system, we expect to reduce the assembly time by assigning the robots an equal number of tasks. Therefore, we have:

$$L_1 = L_2 = L_3 = \frac{L_{wall}}{3} \qquad (2)$$

Another condition that should be satisfied is:

$$L_{wall} \leq n.L_{optimal} \qquad (3)$$

$L_{optimal}$ is the maximum length of wall that each of the robots can cover and build within the optimal time.

### B. Multi-robot placements

The decomposed segment sets($L_1, L_2, L_3$) generated in the path decomposition module, are assigned to the manipulators ($M_1, M_2, M_3$). Since we are using several independent arm robots to collaborate and perform the task, minimizing the assembly time for each of them, leads to a reduction in the overall time required for task completion by the collaborative robotic system. Therefore, in simulation, we focused on finding the best robot's position, material position, and robot configurations for the Panda robot.

## 5. Simulation

We tested our approach with several simulation setups. We defined the robot configuration as 'Ready-Front-01', 'Ready-Front-02', and 'Ready-Side' as the middle stage for IK calculation. The multi-robot placement problem is defined in cartesian coordinates < x, y, z >. Since z is the same in all scenarios, the only two coordinates define the position of the manipulator and material, < x, y >. For each scenario, we load the bricks in the same predetermined location to test the impact of the robot and material position. This means that in one scenario, all bricks possess the same initial positions (Figure 2). We execute the simulation 78 times with four different initial positions of bricks (in 'Ready-Front-01' case five), six target position lines, and three different robot configurations. In the position line parameter, we evaluate the robots' placement effect on brick layering time by changing the distance between the robot and the target line.

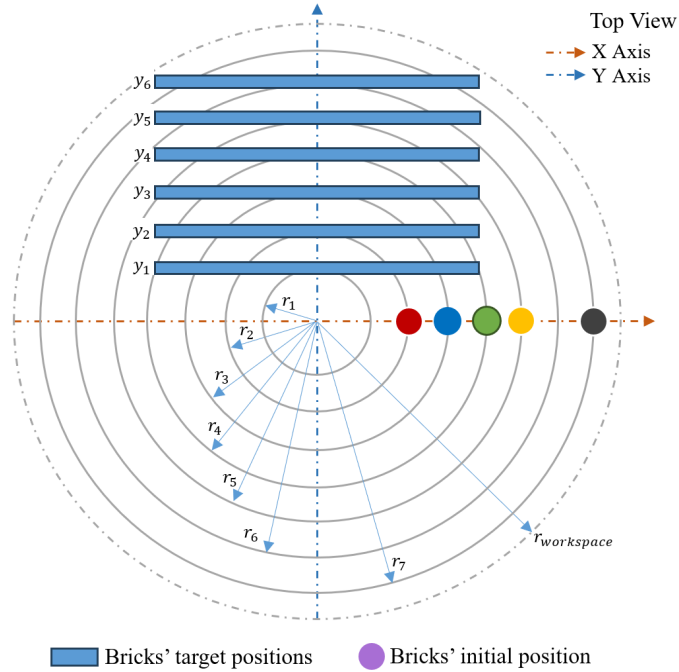

Figure 2: The plot to visualize the robot and material placements with respect to Table 1. parameters

Table 1: Parameters of various simulation scenarios

| Parameters | $r_{workspace}$ | $r_1, y_1$ | $r_2, y_2$ | $r_3, y_3$ | $r_4, y_4$ | $r_5, y_5$ | $r_6, y_6$ | $r_7$ |
|---|---|---|---|---|---|---|---|---|
| Value [m] | 0.85 | 0.2 | 0.3 | 0.4 | 0.5 | 0.6 | 0.7 | 0.8 |

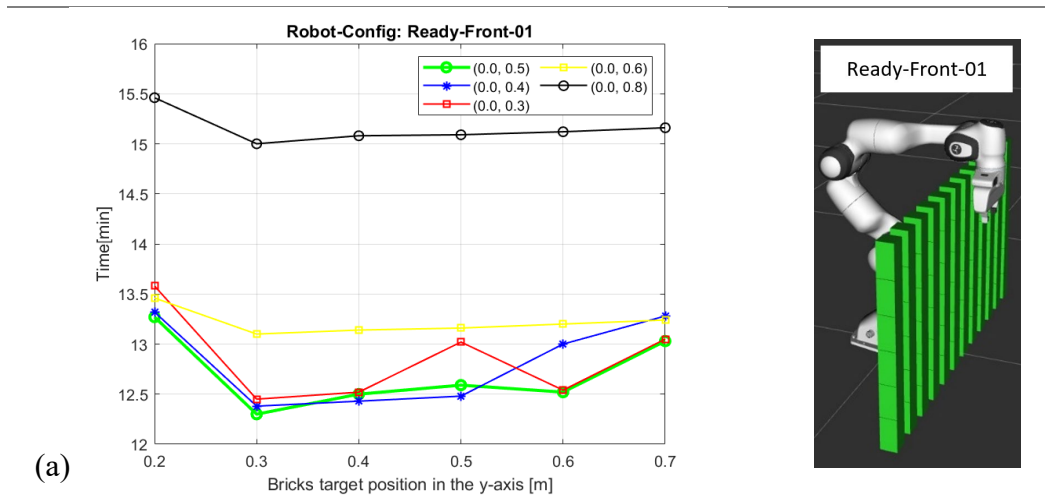

(a)

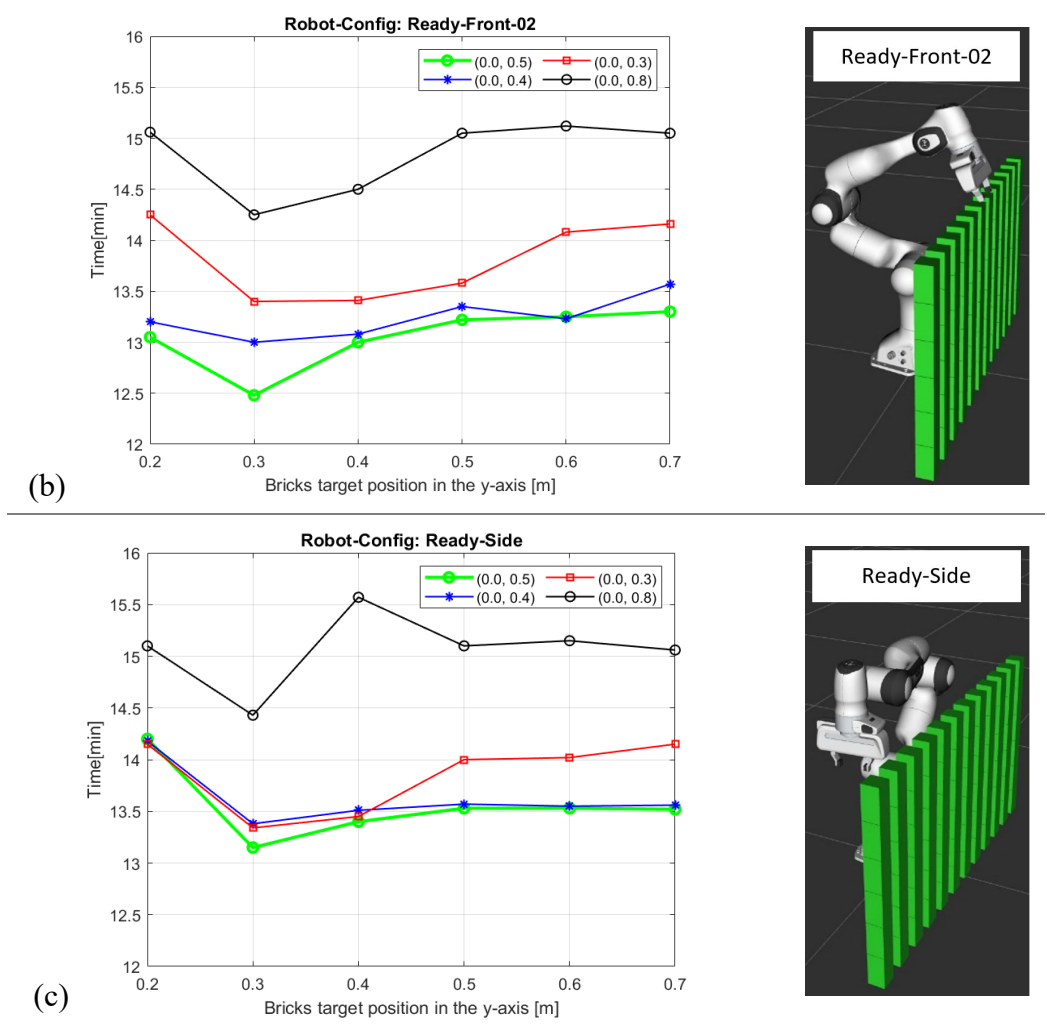

Figure 3: Simulation time result for the Panda robot. (a) assembly time with 'Ready-Front-01' configuration, (b) assembly time with 'Ready-Front-02' configuration, (c) assembly time with 'Ready-Side' configuration

According to the simulation results (Figure 3), in all robot configurations, the time required for bricklaying is minimized when the distance between the robot and the desired wall is $y_2 = 0.3$. Furthermore, when the distance between the target wall and the robot increases, there is a corresponding rise in the duration of bricklaying. Nevertheless, this simulation demonstrates that, in contrast to the idea that a smaller distance between the robot and the wall leads to a quicker construction time, the smallest distance does not necessarily represent the optimal placement. The problem lies in the robot's need to travel a longer path to avoid the collision, which consequently leads to an extended duration for calculating the inverse kinematics. The result identifies an optimal position, $y_2 = 0.3$, where the assembly's duration is minimized.

Moreover, according to the results of the simulation (Figure 3), the optimal initial location of bricks for the Panda robot is determined to be in 0.5m and 0.4m (green and blue circles in Figure 2) on the x-axis. Conversely, despite the idea that the closest material placement is optimal for the initial placement of bricks, the results indicate that $x = 0.3$m does not represent the best material placement. Through the simulation of multiple scenarios, we identified the optimal positioning of bricks and robots. Consequently, we expanded our theory to encompass the coordination of several robots.

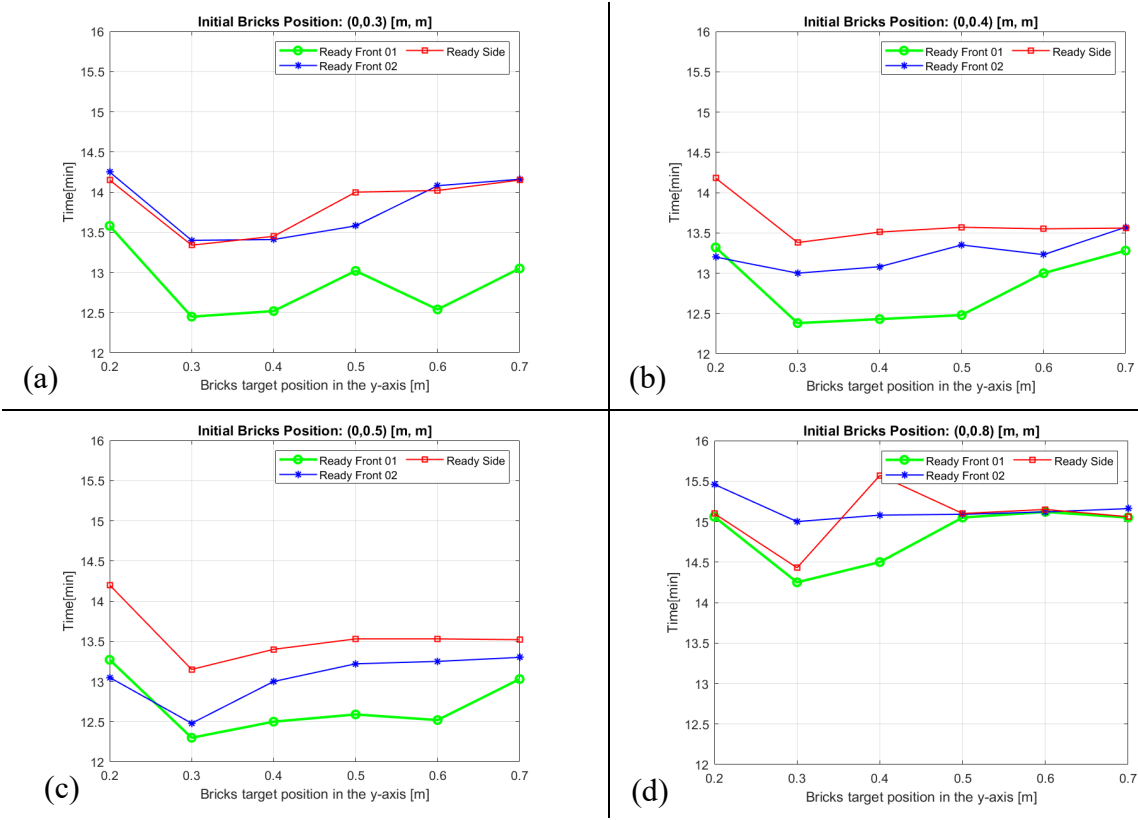

Figure 4: Simulation time results based on different bricks initial position. (a) initial position is 0.3, (b) initial position is 0.4, (c) initial position is 0.5, (d) initial position is 0.8

In Figure 4, we tested the effect of the robot's configuration as the middle stage for IK calculation, with four brick initial placements and six brick target positions. As it is obvious, "Ready-Front-01" is the best configuration among all of them. Finally, based on our simulation, we found the best brick initial placements, robot placements, and robot configuration.

Therefore, we can calculate the $L_{optimal}$, since we obtain the best line of bricks. In our case, the $L_{optimal} = 1.48 \ [m]$ based on equation (4).

$$L_{optimal} = 2 * \sqrt{(r_7{}^2 - r_2{}^2)} \qquad (4)$$

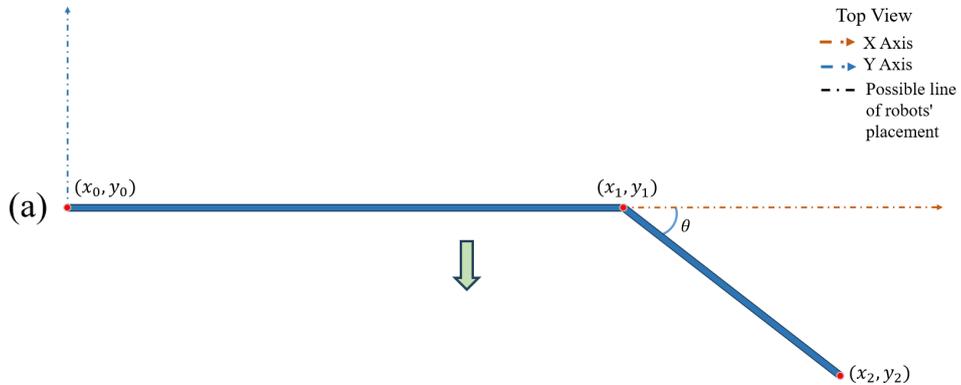

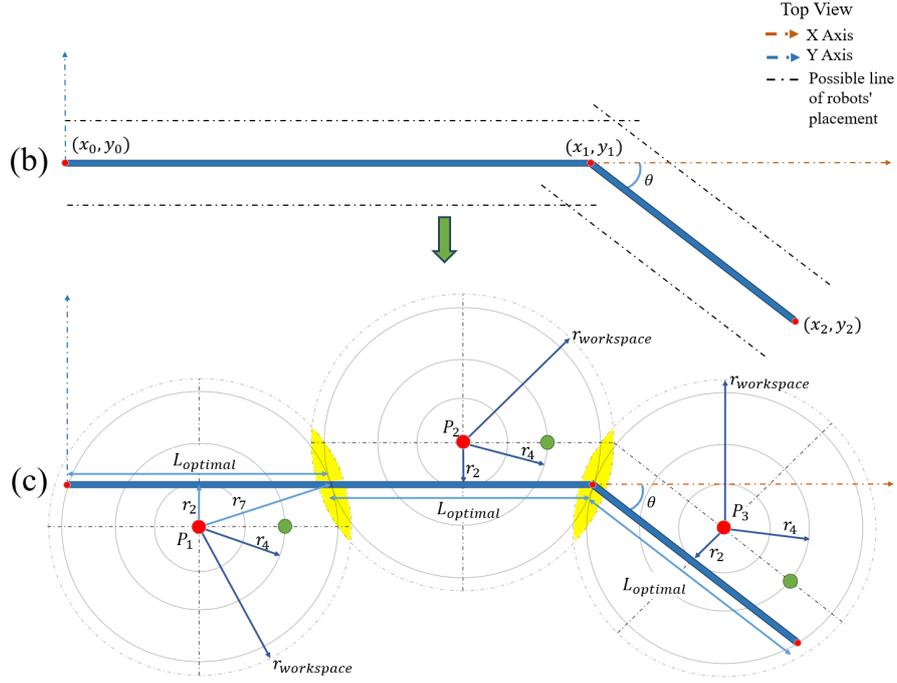

Figure 5: The process of defining robots' placement based on the wall pattern and length. (a) shows the wall pattern and length, (b) shows the lines of robots' positions, (c) shows the multi-Panda coordination and their positions

Finally, our system can suggest the robots' positions based on wall length and pattern.

$$P_1 = (x_0 + \left(\frac{1}{2}\right).L_{optimal}, y_0 - r_2) \tag{5}$$

$$P_2 = (x_0 + \left(\frac{3}{2}\right).L_{optimal}, y_0 + r_2) \tag{6}$$

$$P_3 = (x_0 + \left(2 + \frac{\cos(\theta)}{2}\right).L_{optimal} + r_2\sin(\theta), y_0 - \left(\frac{\sin(\theta)}{2}\right).L_{optimal} + r_2\cos(\theta)) \tag{7}$$

## 6. Conclusion and future steps

This paper presents a novel multi-robot system that provides the best placement for material and robots based on the wall pattern and dimension parameters (Figure 5). We execute simulations to obtain the time for bricklaying with different positions of bricks and robots and robot configurations. The simulation results indicate that $x = 0.5$ and $x = 0.4$ in the x-axis determines the optimal initial position of bricks. Furthermore, the best robot placement is when the distance between the robot and the desired wall is 0.3 ($y_2$). However, the multi-robot placement problem was designed based on a few assumptions that can be addressed in future work. Due to MoveIt's limitations, we could not run the multi-robot simulation simultaneously. Moreover, we enforced the rule allowing only one robot to operate in the shared spaces to prevent collisions between them. Nevertheless, improving the collision constraint is possible to enhance operational efficiency. This paper used the same type of robots; however, future research can extend this by using different arm manipulators.

## Acknowledgement

The presented research was funded by the TUM Innovation Network "CoConstruct" and the DFG Research Unit 5672 "The information backbone of robotized construction" (project no. 517965147).